\documentclass[letterpaper]{article} 
\usepackage{aaai24}  
\usepackage{times}  
\usepackage{helvet}  
\usepackage{courier}  
\usepackage[hyphens]{url}  
\usepackage{graphicx} 
\usepackage{natbib}  
\usepackage{caption} 
\usepackage{algorithm}
\usepackage{algorithmic}

\usepackage{newfloat}
\usepackage{listings}

\usepackage{hhline}
\usepackage{tabularx}
\usepackage{enumitem}
\usepackage{subcaption,siunitx,booktabs}
\usepackage{amsmath}
\usepackage{amssymb}

\DeclareCaptionStyle{ruled}{labelfont=normalfont,labelsep=colon,strut=off} 
\floatstyle{ruled}
\newfloat{listing}{tb}{lst}{}
\floatname{listing}{Listing}
\title{Distribution Matching for  Multi-Task Learning of Classification Tasks:\\ a Large-Scale Study on Faces \& Beyond}
\author{
    Dimitrios Kollias\textsuperscript{\rm 1}\footnote{Corresponding author}, Viktoriia Sharmanska\textsuperscript{\rm 2,\rm 3}, Stefanos Zafeiriou\textsuperscript{\rm 3}
}
\affiliations{
    \textsuperscript{\rm 1} School of Electronic Engineering and Computer Science, Queen Mary University of London, UK\\
     \textsuperscript{\rm 2}School of Engineering and Informatics, University of Sussex, UK \\
     \textsuperscript{\rm 3} Department of Computing, Imperial College London, UK \\
     d.kollias@qmul.ac.uk, sharmanska.v@sussex.ac.uk, s.zafeiriou@imperial.ac.uk
}

\begin{document}

\maketitle

\begin{abstract}
Multi-Task Learning (MTL) is a framework, where multiple related tasks are learned jointly and benefit from a shared representation space, or parameter transfer. 
To provide sufficient learning support, modern MTL uses annotated data with full, or sufficiently large overlap across tasks, i.e., each input sample is annotated for all, or most of the tasks. 
However, collecting such annotations is prohibitive in many real applications, and cannot benefit from datasets available for individual tasks. 
In this work, we challenge this setup and show that MTL can be successful with classification tasks with little, or non-overlapping annotations, or when there is big discrepancy in the size of labeled data per task. 
We explore task-relatedness for co-annotation and co-training, and propose a novel approach, where knowledge exchange is enabled between the tasks via distribution matching. 
To demonstrate the general applicability of our method, we conducted diverse case studies in the domains of affective computing, face recognition, species recognition, and shopping item classification using nine datasets. 
Our large-scale study of affective tasks for basic expression recognition and facial action unit detection illustrates that our approach is network agnostic and brings large performance improvements compared to the state-of-the-art in both tasks and across all studied databases. 
In all case studies, we show that co-training via task-relatedness is advantageous and prevents negative transfer (which occurs when MT model's performance is worse than that of at least one single-task model).
\end{abstract}

\section{Introduction}
\label{sec:intro}

Holistic frameworks, where several learning tasks are interconnected and explicable by the reference to the whole, are common in computer vision. A diverse set of examples includes a scene understanding framework that reasons about 3D object detection, semantic segmentation and depth reconstruction \cite{wang2015holistic}, a face analysis framework that addresses face detection, landmark localization, gender recognition, age estimation \cite{ranjan2017all}, a universal network for low-, mid-, high-level vision \cite{kokkinos2017ubernet}, a large-scale framework of visual tasks for indoor scenes \cite{zamir2018taskonomy}. 
Most if not all prior works rely on building a multi-task framework where learning is done based on the ground truth annotations with full or partial overlap across tasks. During training, all the tasks are optimised simultaneously aiming at representation learning that supports a holistic view of the framework.

What differentiates our work from these holistic approaches is exploring the idea of task-relatedness as means for co-training different tasks. 
In our work, relatedness between  tasks is either provided explicitly in a form of expert knowledge, or is inferred based on empirical studies. 
Importantly, in co-training, the related tasks exchange their predictions and iteratively teach each other so that predictors of all tasks can excel \emph{even if we have limited or no data} for some of them.
We propose an effective distribution matching and co-labeling approach based on distillation \cite{hinton2015distilling}, where knowledge exchange between tasks is enabled via distribution matching over their predictions.

Up until now training holistic models has been primarily addressed by combining multiple datasets to solve individual tasks \cite{ranjan2017all}, or by collecting the annotations in terms of all tasks \cite{zamir2018taskonomy, kokkinos2017ubernet}. 
For example, in affective computing, two most common tasks are predicting categorical expressions (e.g., happy, sad) and activations of binary action units \cite{ekman1997face} (activation of facial muscles) to explain the affective state.
Collecting annotations of AUs is particularly costly, as it requires skilled annotators. 
The datasets collected so far \cite{mollahosseini2017affectnet, EmotioNet2016} have annotations for training only one task and, despite significant effort, there is no dataset that for has complete annotations of both tasks. 
Co-training via task relatedness is an effective way of aggregating knowledge across datasets and transferring it across tasks, especially with little or non-overlapping annotations, or when not many training data are available, or when  there is a big discrepancy in the size of labeled data per task.

In this work we discuss two strategies to infer task-relatedness, via domain knowledge and dataset annotation.  
For example, the two aforementioned tasks of facial behavior analysis are interconnected with known strengths of relatedness in literature. In \cite{ekman1997face}, the facial action coding system (FACS) was built to indicate for each of the basic expressions its \emph{prototypical} AUs. In \cite{du2014compound}, a dedicated user study has been conducted to study the relationship between AUs and  expressions. 
In \cite{khorrami2015deep}, the authors show that DNNs trained for expression recognition implicitly learn AUs. 
In our case study on face recognition, we have a dataset (CelebA \cite{liu2015faceattributes}), where annotations for both tasks, identification and attribute prediction, are available. 
We can infer task relatedness empirically using its annotations.

One of the important challenges in MTL is how to avoid negative transfer, defined as when the performance of the multi-task model is worse than that of at least one single-task model \cite{wang2019characterizing,liu2019loss}. Negative transfer occurs naturally in MTL scenarios when: \\ 
i) source data are heterogeneous or less related (since tasks are diverse to each other, there is no suitable common latent representation and thus MTL produces poor representations); ii) one task or group of related tasks dominates the training process (negative transfer may occur simultaneously on tasks outside the dominant group). 

To overcome negative transfer one can, in the loss function, change the lambdas that control the  importance of some tasks. However: this could severely affect the performance on other tasks; it is a computationally expensive procedure, which lasts many days for each trial; it is an ad-hoc method that does not guarantee to work on other tasks or  databases. To balance the performance on many tasks, \cite{liu2019loss} proposed a method that uses each task's training loss to indicate whether it is well trained, and then decreases the relative weights of the well trained tasks. The evaluation of performance indicators during each training iteration is costly. 
Negative transfer may be induced by conflicting gradients among the different tasks \cite{yu2020gradient}. 
\cite{lin2019pareto} tackled this through multi-objective optimization, with decomposition  of the problem into a set of constrained sub-problems with different trade-off preferences (among different tasks). However, this approach is rather complex, providing a finite set of solutions that do not always satisfy the MTL requirements and finally needs to perform trade-offs among tasks.

We demonstrate empirically that the proposed distribution matching and co-labeling approach based on task relatedness can prevent negative transfer in all our case studies. Via the proposed approach, knowledge of task relationship is infused in network training, providing it, in a simple manner, with higher level representation of the relationship between the tasks;  it is not based on performance indicators and it does not perform any trade-offs between the different tasks.

\noindent The main contributions of this paper are as follows: 
\begin{itemize}[leftmargin=*,noitemsep,nolistsep]

\item We propose a flexible framework that can accommodate different classification tasks by encoding prior knowledge of tasks relatedness. In our experiments we evaluate two effective strategies of task relatedness: 
a) obtained from domain knowledge, e.g., based on  cognitive studies, and b) inferred empirically from dataset annotations (when no domain knowledge is available). 

\item We propose an effective weakly-supervised learning approach that couples, via distribution matching and label co-annotation, tasks with little, or even non-overlapping annotations, or with big discrepancy in their labeled data sizes; we consider a plethora of application scenarios, split in two case studies: i) affective computing; ii) beyond affective computing, including face recognition, fine-grained species categorization, shoe
type classification, clothing categories recognition. 

\item We conduct an extensive experimental study utilizing 9 databases; we show that the proposed method is network agnostic (i.e., it can be incorporated and used in MTL networks) as it brings similar level of performance improvement in all utilized networks, for all tasks and databases. We also show that our method outperforms the state-of-the-art in all tasks and databases. Finally we show that our method successfully prevents negative transfer in MTL.

\end{itemize}

\section{Related Work}

Works exist in literature that use expression labels to complement missing AU annotations or increase generalization of AU classifiers \cite{yang2016multiple,wang2017expression}. Our work deviates from such methods, as we target joint learning of both facial behavior tasks via a single framework, whilst these works perform  only AU detection.
In the face analysis domain, the use of MTL is somewhat limited. 
In \cite{wang2017multi}, MTL was tackled through a  network that jointly handled face recognition and facial attribute prediction tasks. At first a network was trained for facial attribute detection; then it was used to generate attribute labels for another database that only contained face identification labels. Then, a MT network was trained using that database. The network's loss function was the sum of the independent task losses.
In \cite{deng2020multitask}, a unified model performing facial AU detection, expression recognition, and valence-arousal estimation was proposed; the utilized database contained images not annotated for all tasks. To tackle this, authors trained a teacher MT model using the complete labels, then generated pseudo-annotations and finally trained a student MT model  on the union of original and pseudo labels; that network outperformed the teacher one. 
The teacher model did not exploit the tasks being interconnected - the overall loss was equal to the sum of the independent task losses. Thus, the student model did not learn this relatedness. This work utilized only one database.  

In terms of MTL, \cite{sener2018multi} proposed MGDA-UB that casts MTL as multi-objective optimization, with the aim of finding a Pareto optimal solution, and proposed an upper bound for the multi-objective loss. 
\cite{chen2018gradnorm} proposed GradNorm, a gradient optimization algorithm that automatically balances training in MTL by dynamically tuning gradient magnitudes.
\cite{sun2020adashare} proposed AdaShare, an adaptive sharing approach that decides what to share across which tasks by using a task-specific policy optimized  jointly with the network weights.
\cite{strezoski2019many} proposed TR (Task Routing) that applies a conditional
feature-wise transformation over the conv activations and is encapsulated in the conv
layers.

\begin{table*}[t]
\centering
\scalebox{0.87}{
\begin{tabular}{|l|c|c|c|}
\hline
 & \multicolumn{2}{c|}{\begin{tabular}{@{}c@{}}    Cognitive-Psychological  Study \cite{du2014compound} \end{tabular}} & Empirical  Evidences, Aff-Wild2 \\
\hline
expression   & Prototypical AUs & Observational AUs (with weights $w$) & AUs (with weights $w$)\\
\hline\hline
happiness &  12, 25 & 6 (0.51)  & 12 (0.82),\hspace{0.16cm} 25 (0.7),\hspace{0.16cm} 6 (0.57),\hspace{0.16cm} 7 (0.83),\hspace{0.16cm} 10 (0.63) \\
\hline
sadness &  4, 15 & 1 (0.6),\hspace{0.16cm} 6 (0.5),\hspace{0.16cm} 11 (0.26),\hspace{0.16cm} 17 (0.67) & 4 (0.53),\hspace{0.16cm} 15 (0.42),\hspace{0.16cm} 1 (0.31),\hspace{0.16cm} 7 (0.13) \\
\hline
fear &  1, 4, 20, 25 &2 (0.57),\hspace{0.16cm} 5 (0.63),\hspace{0.16cm} 26 (0.33) & 1 (0.52),\hspace{0.16cm} 4 (0.4),\hspace{0.16cm} 25 (0.85),\hspace{0.16cm} 7 (0.57),\hspace{0.16cm} 10 (0.57) \\
\hline
anger &4, 7, 24 &10 (0.26),\hspace{0.16cm} 17 (0.52),\hspace{0.16cm} 23 (0.29) & 4 (0.65),\hspace{0.16cm} 7 (0.45),\hspace{0.16cm} 25 (0.4),\hspace{0.16cm} 10 (0.33)\\
\hline
surprise &1, 2, 25, 26 &5 (0.66)  & 1 (0.38),\hspace{0.16cm} 2 (0.37),\hspace{0.16cm} 25 (0.85),\hspace{0.16cm} 26 (0.3),\hspace{0.16cm} 5 (0.5),\hspace{0.16cm} 7 (0.2) \\
\hline
disgust &9, 10, 17 & 4 (0.31),  \hspace{0.16cm} 24 (0.26) &  10 (0.85), \hspace{0.16cm} 4 (0.6), \hspace{0.16cm} 7 (0.75), \hspace{0.16cm} 25 (0.8)\\
\hline
\end{tabular}
}
\caption{Relatedness of expressions \& AUs inferred from \cite{du2014compound} (the weights denote fraction of annotators that observed the AU activation) or from Aff-Wild2 (the weights denote percentage of images that the AU was activated)}
\label{table:EmoAUs}
\end{table*}

\section{The Proposed Approach}\label{approach}

Let us consider a set of \textit{m} classification tasks $ {\{ \mathcal{T}_i \}}^m_{i=1}$. 
In task $\mathcal{T}_i$, the observations are generated by the underlying distribution $\mathcal{D}_i$ over inputs $\mathcal{X}$ and their labels $\mathcal{Y}$ associated with the task. 
For the \textit{i}-th task $T_i$, the training set ${D}_i$ consists of $n_i$ data points $ (\mathbf{x}_j^i, y_j^i),$ $j= 1,\ldots,n_i$ with $\mathbf{x}_j^i \in \mathbb{R}^d$ and its corresponding output $y_j^i \in \{0,1\}$ if it is a binary classification task, or $y_j^i \in \{0,1\}^k$ (one-hot encoding) if it is a (mutually exclusive) k-class classification task.

The goal of MTL is to find \textit{m} hypothesis: $h_1, . . . , h_m$ over the hypothesis space $\mathcal{H}$ to control the average expected error over all tasks: $ \frac{1}{m} \sum_{i=1}^{m}{\mathbb{E}_{(\mathbf{x},y)  \sim \mathcal{D}_i}} \mathcal{L}(h_i(\mathbf{x}), y)$ with $\mathcal{L}$ being the loss function. We can also define a weight $\mathbf{w_i} \in \Delta^m$, ${\{ \mathbf{{w}}_{i} \}}^m_{i=1} > 0$ to govern each task's contribution. The overall loss is: 
$
 \mathcal{L}_{MT} = \frac{1}{m} \sum_{i=1}^{m}{ \mathbf{w_i} \cdot \mathbb{E}_{(\mathbf{x},y)  \sim \mathcal{D}_i}} \mathcal{L}(h_i(\mathbf{x}),y) $.

In the following, we present the proposed framework via a plethora of case studies, mainly focusing on  affective computing. The framework includes inferring the tasks' relationship and using it for coupling them during MTL. The coupling  is achieved via the proposed co-annotation and distribution matching losses, which can be incorporated and used in any network that performs MTL, regardless of the input modality (visual, audio, text). The advantages of using these  losses include: i) flexibility: no changes to network structure are made and  no additional burden on inference is placed; ii) effectiveness: performance of various networks on multiple databases (small- or large-scale, image or video) is boosted and negative transfer is alleviated; iii) efficiency: negligible computational complexity is added during training; iv) easiness: a few lines of code are needed to implement.

\subsection{Case Study I: Affective Computing}

We start with the multi-task formulation of the  behavior model. 
In this model we have two objectives: (1) learning 7 basic expressions, 
(2) detecting activations of $17$ binary AUs. 
We train a multi-task model to jointly perform (1)-(2). 
For a given image $x \in \mathcal{X}$, we can have label annotations of either one of 7 basic expressions $y_{exp} \in \{1,2,\ldots,7\}$, or $M$ AU activations $y_{au} \in \{0,1\}^{M}$.
For simplicity of presentation, we use the same notation $x$ for all images leaving the context to be explained by the label notations.  
We train the multi-task model by minimizing the following objective:

\begin{equation}
\mathcal{L}_{MT} =  \mathcal{L}_{Exp} +  \mathcal{L}_{AU} +  \mathcal{L}_{DM} +  \mathcal{L}_{SCA}  
\end{equation}

\noindent where: $\mathcal{L}_{Exp}=\mathbb{E}_{x,y_{exp}}\big[-\text{log } p (y_{exp}|x)\big]\nonumber$ is cross entropy (CE) loss computed over images with basic emotion label; $\mathcal{L}_{AU}=\mathbb{E}_{x,y_{au}}\big[- \text{log } p (y_{au}|x)\big]$ is binary CE loss computed over images with $M$ AU activations, with: $\text{log } p(y_{au}|x) := $ 

\noindent $\frac{\sum_{i=1}^{M} \delta_i \cdot [y_{au}^i\text{log } p (y_{au}^i|x) + (1-y_{au}^i)\text{log } (1-p (y_{au}^i|x))]} {\sum_{k=1}^{M} \delta_k}$, $\delta_i \in \{0,1\}$ indicates if the image contains $AU_i$ annotation; $\mathcal{L}_{DM}$ and $\mathcal{L}_{SCA}$ are the distribution matching and soft co-annotation losses (i.e., the proposed coupling losses) based on the relatedness between expressions and AUs; the losses' derivation is explained in the following.

\subsubsection{\textbf{Task-Relatedness} \\} 

\underline{1) Obtained from Domain Knowledge:} In the seminal work of \cite{du2014compound}, a cognitive and psychological study on the relationship between expressions and facial AU activations is conducted. The summary of the study is a Table of the relatedness between expressions and their prototypical and observational AUs, that we include in Table \ref{table:EmoAUs} for completeness. Prototypical AUs are ones that are labelled as activated across all annotators' responses; observational are AUs that are labelled as activated by a fraction of annotators.

\noindent
\underline{2) Inferred Empirically from Dataset Annotations:} If the above  cognitive study is not available, we can infer task relatedness from external dataset annotations. In particular, we use the training set of Aff-Wild2 database \cite{kollias2019expression,kollias2021affect,kollias2021analysing,kollias2017recognition,kollias2019deep,kollias2020analysing,kollias2023cvpr,zafeiriou2017aff} to infer task relatedness, since this dataset is the first in-the-wild one that is fully annotated with basic expressions and AUs; this is shown in 
Table \ref{table:EmoAUs}. 
\noindent In the following, we use as domain knowledge the cognitive and psychological study \cite{du2014compound}, to encode task relatedness and introduce the proposed approach for coupling the tasks.

\subsubsection{\textbf{Coupling of basic expressions and AUs}}  \label{distr_matching2}

\noindent\paragraph{Via Distribution Matching}
Here, we propose the distribution matching loss for coupling the expression and AU tasks.
The aim is to align the \emph{predictions} of expression and AU tasks during training by making them consistent.
From expression predictions we create new soft AU predictions and then match these with the network's actual AU predictions. For instance, if the network predicts \emph{happy} with probability 1 and also predicts that AUs 4, 15 and 1 are activated, this is a mistake as these AUs are  associated with the expression \emph{sad} according to the prior knowledge. With this loss we infuse the prior knowledge into the network to guide the generation of better and consistent predictions.

For each sample $x$ we have the predictions of expressions $p(y_{exp}|x)$ as the softmax scores over seven basic expressions and we have the prediction of AUs activations $p(y_{au}^i|x)$, $i=1,\ldots,M$ as the sigmoid scores over $M$ AUs. 
We  match the distribution over AU predictions $p(y_{au}^i|x)$ with the distribution 
$q(y_{au}^i|x)$, where the AUs are modeled as a mixture over the basic expression categories: 
\begin{equation}
    q(y_{au}^i|x) = \sum_{y_{exp} \in \{1,\ldots,7\}} p(y_{exp}|x) \: p(y_{au}^i| y_{exp}), 
\label{eq:distr}
\end{equation} 
where $p(y_{au}^i| y_{exp})$ is defined deterministically from Table~\ref{table:EmoAUs} 
and is 1 for prototypical/observational action units, or 0 otherwise.  For example, AU2 is prototypical for expression \emph{surprise} and observational for expression \emph{fear} and thus $q(y_{\text{AU2}}|x) = p(y_{\text{surprise}}|x) + p(y_{\text{fear}}|x)$. 
So with this matching if, e.g., the network predicts the expression \emph{happy} with probability 1, i.e., $p(y_{\text{happy}}|x)=1$, then the prototypical and observational AUs of \emph{happy} -AUs 12, 25 and 6- need to be activated in the distribution q: $q(y_{\text{AU12}}|x) = 1$; $q(y_{\text{AU25}}|x) = 1$; $q(y_{\text{AU6}}|x) = 1$; $q(y_{au}^i|x) = 0$, $i \in \{1,..,14\}$. 

In spirit of the distillation approach, we match the distributions $p(y_{au}^i|x)$ and $q(y_{au}^i|x)$ 
by minimizing the cross entropy with the soft targets loss term: 
\begin{align}
\mathcal{L}_{DM} = \mathbb{E}_{x} \Bigg[ \sum_{i=1}^{M}[ -p(y_{au}^i|x)\text{log }q(y_{au}^i|x)] \Bigg] , \label{eq:coupleloss}
\end{align}
where all available train samples are used to match the predictions. \\

\noindent\textbf{Via soft  co-annotation} 
Here, we propose the soft co-annotation loss for coupling the expression and AU tasks.
At first we create soft expression labels (that are guided by AU labels) by infusing prior knowledge of their relationship. Then we match these labels with the expression predictions.
The new expression labels will help in cases of images with partial or no annotation overlap, especially if there are not many training data.  
We use the AU labels (instead of predictions) as they provide more confidence (the AU predictions -especially at the beginning of training- will be quite wrong; if we utilized this loss with the wrong AU predictions, it would also affect negatively the expression predictions).

Given an image $x$ with ground truth AU annotations, $y_{au}$, we first co-annotate it with a \emph{soft label} in form of the  distribution over expressions and then match it with the predictions of expressions $p(y_{exp}|x)$.
Thus, at first we compute, for each basic expression, an indicator score, $I(y_{exp}|x)$ over its prototypical and observational AUs being present: 

\begin{align}
    I(y_{exp}|x) = \frac{ \sum_{ i \in \{1,\ldots,M\}} w_{au}^i \cdot y_{au}^i} {\sum_{i \in \{1,\ldots,M\}} w_{au}^i} , \text{ $y_{exp} \in \{1,\ldots,7\}$} 
\end{align} 

\noindent where: $w_{au}^i$ is 1 if AU $i$ is prototypical for $y_{exp}$ (from Table \ref{table:EmoAUs}); is $w$ if AU $i$ is observational for $y_{exp}$; is 0 otherwise.

\noindent For example, for expression \emph{happy}, the indicator score $I(happy|x) = (y_{\text{AU12}} + y_{\text{AU25}} + 0.51 \cdot y_{\text{AU6}}) / (1+1+0.51)$.

Then, we convert the indicator scores to probability scores over expression categories; this \emph{soft} expression label, $q(y_{exp}|x)$, is computed as following:

\begin{align}
    q(y_{exp}|x) &=  \frac{e^{I(y_{exp}|x)}}{\sum_{y'_{exp}} e^{I(y'_{exp}|x)}}
     , \text{ $ \{y_{exp}, y'_{exp}\}  \in \{1, ..,7\}$}
\end{align}

In this variant, every single image that has ground truth annotation of AUs will have a \emph{soft} expression label assigned. Finally we match the predictions $p(y_{exp}|x)$ and the \emph{soft} expression label $q(y_{exp}|x)$ by minimizing the cross entropy with the soft targets loss term:

\begin{align}
\mathcal{L}_{SCA} =  \mathbb{E}_{x} \Bigg[ \sum_{y_{exp}\in \{1,\ldots,7\}}[ -p(y_{exp}|x)\text{log }q(y_{exp}|x)] \Bigg] \label{eq:coupleloss2}
\end{align}

\subsection{Case Study II: Beyond Affective Computing}

Here, we show that our approach can also be used in other application scenarios: i) face recognition (facial attribute detection and face identification); ii) fine-grained species categorization (species classification and attribute detection); iii) shoe type recognition (shoe type classification and attribute detection); iv) clothing categories recognition (classification of clothing categories and attributes).

In the model's multi-task (MT) formulation, we have two objectives: (1) to detect $M$ binary attributes, 
 (2) to classify $N$ classes. 
The aim of a MT model is to jointly perform (1) and (2). 
For a given image $x \in \mathcal{X}$, we can have labels of one of $N$ classes $y_{cls} \in \{1,\ldots,N\}$, and $M$ binary attributes $y_{att} \in \{0,1\}^{M}$.
We train the MT model by minimizing the objective: $\mathcal{L}_{MT} = \mathcal{L}_{Clc} +  \mathcal{L}_{Att} +  \mathcal{L}_{DM} +  \mathcal{L}_{SCA}$, where:
\begin{equation}
{L}_{DM} = \mathbb{E}_{x} \Bigg[ \sum_{i=1}^{M}[ -p(y_{att}^i|x)\text{log}
\sum_{y_{cls}} p(y_{cls}|x) \: p(y_{att}^i| y_{cls}) 
] 
\Bigg] \nonumber  
\end{equation}
\begin{equation}
\mathcal{L}_{SCA} =  \mathbb{E}_{x} \Bigg[ \sum_{y_{cls}}[ -p(y_{cls}|x)\text{log }
\frac{e^{I(y_{cls}|x)}}{\sum_{y'_{cls}} e^{I(y'_{cls}|x)}}
] \Bigg] \nonumber \\ \nonumber \\ \nonumber 
\end{equation}
\noindent
$\mathcal{L}_{Cls}$ is the cross entropy loss for the classification task; \\
$\mathcal{L}_{att}$ is the binary cross entropy loss for the detection task;
$\mathcal{L}_{DM}$ is the distribution matching loss for matching the distributions $p(y_{att}^i|x)$ and the one where the attributes are modeled as a mixture over the classes; 
\\
$\mathcal{L}_{SCA}$ is the soft co-annotation loss for matching predictions $p(y_{cls}|x)$ and \emph{soft} class labels (i.e., probability of each class indicator score, $I(y_{cls}|x)$, over its detected attributes); 
\\
$p(y_{att}^i| y_{cls}) = \frac{\text{total number of images with both } y_{att}^i \text{ and } y_{cls}}{\text{total number of images with } y_{cls}} $, is inferred empirically from dataset annotations.

\begin{table*}[ht]
\centering
\scalebox{0.78}{
\begin{tabular}{ |c||c|c|c|c|c|c|c| }
 \hline
\multicolumn{1}{|c||}{\begin{tabular}{@{}c@{}} Databases  \end{tabular}}  &  \multicolumn{2}{c|}{\begin{tabular}{@{}c@{}}  AffectNet - EmotioNet \end{tabular}} & \multicolumn{2}{c|}{\begin{tabular}{@{}c@{}}  RAF-DB - EmotioNet \end{tabular}} & \multicolumn{2}{c|}{\begin{tabular}{@{}c@{}}  ABAW4 LSD - EmotioNet\end{tabular}}  &  \multicolumn{1}{c|}{\begin{tabular}{@{}c@{}}  Aff-Wild2 \end{tabular}}   \\
 \hline
Methods & EmoAffectNet & 	
 EffNet-B2  & 
PSR  & VGGFACE  &    MTER-KDTD & HSE-NN &  TMIF-FEA\\
 \hline
Metrics &  Acc - AFA   & Acc - AFA & AA - AFA  &  AA - AFA & F1 - AFA & F1 - AFA &  F1 (Emo) - F1 (AU)\\ 
  \hline
 \hline

ST   &  0.664 - 0.719$^\star$  &  0.663 - 0.705$^\star$  &  0.808 - 0.696$^\star$ & 0.775 - 0.693$^\star$ & 0.359 - 0.70$^\star$   &  0.372 - 0.708$^\star$    & 0.359 - 0.499   \\
\hline
\hline

\begin{tabular}{@{}c@{}}  NC MT \end{tabular}  &   0.643$^\star$ - 0.751$^\star$ & 0.639$^\star$ - 0.738$^\star$  & 0.786$^\star$ - 0.728$^\star$  &  0.754$^\star$ - 0.722$^\star$ & 0.342$^\star$ - 0.722$^\star$ & 0.354 - 0.73$^\star$  & 0.337$^\star$ - 0.52$^\star$ \\
\hline
\hline

\begin{tabular}{@{}c@{}}  S-T NC MT \end{tabular}  &   0.649$^\star$ - 0.761$^\star$ & 0.645$^\star$ - 0.749$^\star$  & 0.798$^\star$ - 0.732$^\star$  &  0.765$^\star$ - 0.726$^\star$ & 0.35$^\star$ - 0.729$^\star$ & 0.361 - 0.737$^\star$ & 0.339$^\star$ - 0.522$^\star$ \\
\hline
\hline

\begin{tabular}{@{}c@{}} \textbf{C MT (DM)}   \end{tabular} &   \textbf{0.694} - \textbf{0.80} & \textbf{0.689} - 0.785 &   \textbf{0.848} - \textbf{0.781}  &  \textbf{0.814} - \textbf{0.772} & 0.386  - 0.775 & 0.393 - 0.783 & 0.388 - 0.559 \\
\hline

\begin{tabular}{@{}c@{}} \textbf{C MT}  \textbf{ (Aff-Wild2)}   \end{tabular}  &   \textbf{0.694} - \textbf{0.80} & 0.687 - \textbf{0.787} & 0.838 - 0.772  & 0.805 - 0.76  & \textbf{0.395} - \textbf{0.785} & \textbf{0.403} - \textbf{0.792} &  \textbf{0.399} - \textbf{0.578} \\
\hline
\end{tabular}
}
\caption{Performance comparison between various state-of-the-art single-task (ST) methods vs their multi-task counterparts with/without coupling (C MT/NC MT, respectively) under two relatedness scenarios (DM or Aff-Wild2) vs their Student-Teacher (S-T) knowledge distillation counterparts; 
$^\star$ denotes our own implementation; 'Acc': Accuracy; 'AA': Average Acc.; 'AFA': average between F1 and Acc.;  DM is domain knowledge from \cite{du2014compound}} 
\label{comparison_sota}
\end{table*}

\section{Experimental Study}

\paragraph{Databases} \label{databases}

In this work we utilized:
\textbf{AffectNet} \cite{mollahosseini2017affectnet} with around  350K in-the-wild images annotated for 7 basic expressions;
\textbf{RAF-DB} \cite{li2017reliable} with around 15K in-the-wild images annotated for 7 basic expressions;
%
\textbf{ABAW4 LSD} \cite{kollias2022eccv}  -utilized in 4th Affective Behavior Analysis in-the-wild (ABAW) Competition at ECCV 2022- with around 280K in-the-wild synthetic images and 100K in-the-wild real images (which constitute the test set) annotated for 6 basic expressions;  
%
\textbf{Aff-Wild2} \cite{kollias2022cvpr} -as utilized in 3rd ABAW Competition at CVPR 2022- with 564 in-the-wild videos (A/V) of around 2.8M frames annotated for 7 basic expressions (plus 'other'), 12 AUs and valence-arousal;
\textbf{EmotioNet} \cite{fabian2016EmotioNet} with around 50K images manually annotated for 11 AUs;
\textbf{CelebA} with around 205K in-the-wild images of around 10.2K identities, each with 40 attributes (its training, validation and test sets are subject independent; for our experiments, we generated a new split into 3 subject dependent sets); 
\noindent
\textbf{Caltech-UCSD Birds-200-2011} \cite{WahCUB_200_2011} (CUB) with around 12K images of 200 bird species and of 312 binary attributes;
\textbf{Shoes} \cite{WahCUB_200_2011} (S-ADD) with around 15K women shoe images of 10 different types and of 10 attributes \cite{kovashka2012whittlesearch} (it does not contain any predefined split, thus we split it in a training set of 7.4K and a test set of 7.3K images).
\textbf{Clothing Attributes Dataset} \cite{chen2012describing} (CAD) with around 2K images partially annotated for 7 clothing categories, 23 binary and 2 multi-class attributes (due to its very small size and to the non-predefined split, we perform 6 times 2-fold cross validation, i.e., we create 6 different 50-50 splits of the data).

\paragraph{Performance Measures}
We use: 
i) average accuracy (AA) for RAF-DB; ii) accuracy for AffectNet; iii) the average between F1 and mean accuracy for EmotioNet (AFA); iv) F1 for ABAW4 LSD and Aff-Wild2; vii) total accuracy and F1 for CelebA, CUB, S-ADD, CAD.

\paragraph{Pre-Processing \& Training Implementation Details}

Case Study I: We used RetinaFace  \cite{deng2020retinaface} to extract bboxes  and 5 facial landmarks (for alignment) and resized images   to $112 \times 112 \times 3$. Mixaugment \cite{psaroudakis2022mixaugment} was used for data augmentation. 
Case Study II: for CelebA we used the database's aligned images and resized them to $112 \times 112 \times 3$ (batch size = 200, Adam, lr = $10^{-3}$); for CUB, we cropped the images using the bboxes, resized them to $280 \times 280 \times 3$, we used label smoothing (value = 0.3) and performed  affine transformations as data augmentation (batch size = 150, Adam, lr = $10^{-3}$); for S-ADD, we resized images to $280 \times 280 \times 3$ (batch size = 100, Adam, lr = $10^{-3}$); for CAD, we resized images  to $280 \times 280 \times 3$ (batch size = 50, Adam, lr = $10^{-3}$). In all databases, images were  normalized  to  $[-1,1]$. 
Tesla V100 32GB GPU \& Tensorflow were used for model training.

\subsection{Results on Case Study I: Affective Computing}

\paragraph{Effectiveness of proposed coupling losses across various networks}

We utilized the state-of-the-art for databases: i) AffectNet: EmoAffectNet\cite{ryumina2022search} and EffNet-B2 \cite{savchenko2021facial}; ii) RAF-DB: PSR \cite{vo2020pyramid} and VGGFACE \cite{kollias2020deep}; iii) ABAW4 LSD: HSE-NN \cite{savchenko2022hse} and MTER-KDTD \cite{jeong2022learning}; 
iv)  Aff-Wild2: TMIF-FEA \cite{zhang2022transformer}. 
Let us note that EffNet-B2, MTER-KDTD and HSE-NN are multi-task methods.

The result of using each state-of-the-art (sota) in single-task manner is shown in the row 'ST' (i.e., Single Task) of Table \ref{comparison_sota}.  
The result of using each sota in MTL manner (e.g.,  EmoAffectNet trained on both AffectNet and EmotioNet;  PSR trained on both RAF-DB and EmotioNet) is shown in  row 'NC MT' (i.e., Multi-Task without coupling) of Table \ref{comparison_sota}. 
It might be argued that since more data are used for network training (i.e., the  additional data coming from multiple tasks, even if they contain partial or non-overlapping annotations), the MTL performance will be better for all tasks. However, as shown and explained next, this is not the case as negative transfer can occur, or sub-optimal models can be produced for some, or even all tasks \cite{wu2019understanding}.

It can be seen in  Table \ref{comparison_sota} (rows 'ST' and 'NC MT'), for all databases, that  the sota,  when trained in a MTL manner (without coupling),  display a better performance for AU detection, but an inferior one for expression recognition - when compared to the corresponding performance of the single-task sota. \textit{This indicates that negative transfer occurs in the case of basic expressions}. This negative transfer effect was due to the fact that the AU detection task dominated the training process. In fact, the EmotioNet database has a larger size than the RAF-DB, AffectNet and ABAW4 LSD. Negative transfer largely depends on the size of labeled data per task \cite{wang2019characterizing}, which has  a direct effect on the feasibility and reliability of discovering shared regularities between the joint distributions of the tasks in MTL. 
 
Finally, we trained each of the sota networks in a MTL manner with the proposed coupling, under two  relatedness scenarios; when the relatedness between the expressions and AUs was derived from the cognitive and psychological study of \cite{du2014compound}, or from dataset annotations (from Aff-Wild2 database). The former case is shown in  row 'C MT (DM))' of Table \ref{comparison_sota} and the latter case  in row 'C MT (Aff-Wild2)'. From Table \ref{comparison_sota} two observations can be made. 

\underline{Firstly}, when the proposed coupling  is conducted, in each sota multi-task network, negative transfer is alleviated; the performance of all multi-task networks is better than the corresponding one of the single-task counterparts for both tasks. This is consistently observed in all utilized databases and experiments.
\underline{Secondly}, the use of the proposed coupling  brings similar levels of performance improvement in all sota multi-task networks across the databases. In more detail, when coupling is conducted, networks outperform their counterparts without coupling by approximately: i) 5\% on Affectnet and 5\% on EmotioNet (both EmoAffectNet and EffNet-B2); ii) 6\% on RAF-DB and 5\% on EmotioNet (both PSR and VGGFACE); iii) 5\% on ABAW4 LSD and 6\% on EmotioNet (both MTER-KDTD and HSE-NN); iv) 5.5\% on Aff-Wild2 (MTER-KDTD).

To sum up, the use of coupling makes the MT networks greatly outperform their MT (without coupling) and single-task counterparts. This proves that the proposed coupling losses are network and modality agnostic as they can be applied and be effective in different networks and different modalities (visual, audio, A/V and text; e.g. TMIF-FEA is a multi-modal approach).  
This stands no matter which task relatedness scenario has been used for coupling the two tasks. 

Finally, for comparison purposes we also used the Student-Teacher (S-T) knowledge distillation approach. We used one, or multiple teacher networks to create soft-labels for the databases that contain annotations only for one task, so that they contain complete, overlapping annotations for both tasks; we then trained a multi-task network on them. 
To illustrate this via an example: we use the single-task EmoAffectNet trained on AffectNet for expression recognition and test it on EmotioNet to create soft-expression labels; thus EmotioNet contains its  AU labels and soft expression labels, i.e., the predictions of EmoAffectNet. Then we use the single-task EmoAffectNet trained on EmotioNet for AU detection and test it on AffectNet to create soft-AU labels; thus AffectNet contains its expression labels and soft AU labels. Then we train EmoAffectNet for MTL using both databases. We compare its performance to EmoAffectNet trained for MTL with the proposed coupling losses on both databases with their original non-overlapping annotations.

The results of the S-T approach are denoted in
row ’S-T NC MT’ (denoting Student-Teacher Multi-Task with no coupling) of Table \ref{comparison_sota}. It can be observed that this approach shows a slightly better performance in both tasks compared to the multi-task counterparts that have been trained with the original non-overlapping annotations -without coupling-. This is expected as these networks have been trained with more annotations for both tasks. \textit{Nevertheless, negative transfer for the basic expressions still occurs}. Moreover, our proposed approach greatly outperforms the S-T one. So overall, our proposed approach alleviates negative transfer and also brings bigger performance gain than that of S-T  approach.

\begin{table}[ht]
\centering
\scalebox{0.84}{
\begin{tabular}{ |c||c||c|c|c|c|c|c|c|c|c|c| }
 \hline
\multicolumn{1}{|c||}{\begin{tabular}{@{}c@{}}   TMIF-FEA\end{tabular}} & \multicolumn{1}{c||}{\begin{tabular}{@{}c@{}} Relatedness \end{tabular}}  & \multicolumn{2}{c|}{Aff-Wild2} \\
 \hline
 &   & F1 (expression) & F1 (AU)
  \\ 
  \hline \hline
no coupling & -  & 0.337 & 0.52   \\
 \hline
 \hline
soft co-annotation & DM  & 0.377 & 0.544  \\
 \hline
 distr-matching & DM   & 0.374  & 0.547   \\ 
 \hline 
 \textbf{both}  & DM & 0.388 & 0.559      \\
 \hline 
\hline
soft co-annotation & Aff-Wild2  & 0.388 & 0.561  \\
 \hline
 distr-matching & Aff-Wild2  & 0.383  & 0.565   \\ 
 \hline 
\begin{tabular}{@{}c@{}} \textbf{both}   \end{tabular} &  Aff-Wild2 &  \textbf{0.399}   & \textbf{0.578}   \\ 
 \hline
\end{tabular}
}
\caption{Ablation Study on TMIF-FEA  with/without coupling, under two task relatedness scenarios; DM is domain knowledge from \cite{du2014compound}}
\label{comparison_losses}
\end{table}

\begin{table*}[ht]
\centering
\scalebox{0.85}{

\begin{tabular}{ |c||c||c|c||c|c||c|c||c|c||c|c||c|c|| }
 \hline
\multicolumn{1}{|c||}{Network} & \multicolumn{1}{c||}{Database} & \multicolumn{12}{c||}{Setting}   \\
 \hhline{|=|=|============|}
  & & 
 \multicolumn{4}{c||}{\begin{tabular}{@{}c@{}}  $2 \times$ ST \end{tabular}} & 
 \multicolumn{4}{c||}{\begin{tabular}{@{}c@{}} NC MT \end{tabular}} &
 \multicolumn{4}{c||}{\begin{tabular}{@{}c@{}}  C MT \end{tabular}}  \\
  \hline
  & & 
   \multicolumn{2}{c||}{Classes} &  \multicolumn{2}{c||}{Attributes} &
 \multicolumn{2}{c||}{Classes} &  \multicolumn{2}{c||}{Attributes} &
  \multicolumn{2}{c||}{Classes} &  \multicolumn{2}{c||}{Attributes} \\
    \hline
  & & 
   \multicolumn{1}{c|}{Acc} &  \multicolumn{1}{c||}{F1} & \multicolumn{1}{c|}{Acc} &  \multicolumn{1}{c||}{F1} &
 \multicolumn{1}{c|}{Acc} &  \multicolumn{1}{c||}{F1} & \multicolumn{1}{c|}{Acc} &  \multicolumn{1}{c||}{F1} &
  \multicolumn{1}{c|}{Acc} &  \multicolumn{1}{c||}{F1} & \multicolumn{1}{c|}{Acc} &  \multicolumn{1}{c||}{F1}  \\
  \hline \hline
  
\begin{tabular}{@{}c@{}} VGG \end{tabular} & \begin{tabular}{@{}c@{}} CelebA \\ CUB \\ S-ADD \\ CAD  \end{tabular}  &
\begin{tabular}{@{}c@{}}  78.11 \\ 78.23 \\ 71.87 \\  52$\pm$5 \end{tabular}  & 
\begin{tabular}{@{}c@{}} 70.02 \\ 78.44 \\ 71.24  \\  38$\pm$7 \end{tabular} &
 \begin{tabular}{@{}c@{}} 87.57 \\ 85.18 \\ 91.05 \\  80$\pm$2 \end{tabular} & 
\begin{tabular}{@{}c@{}} 67.88 \\ 27.01 \\ 89.23 \\  40$\pm$2   \end{tabular} & 
\begin{tabular}{@{}c@{}} 80.75 \\ 80.02 \\ 72.21 \\  41$\pm$7 \end{tabular} & 
\begin{tabular}{@{}c@{}} 71.98 \\ 80.18 \\ 72.01 \\  32$\pm$9 \end{tabular} &  
\begin{tabular}{@{}c@{}} 89.39 \\ 85.54 \\ 90.44 \\  75$\pm$3 \end{tabular} & 
\begin{tabular}{@{}c@{}} 68.65 \\ 28.59 \\ 88.51 \\  33$\pm$4  \end{tabular} & 
\begin{tabular}{@{}c@{}} \textbf{84.98} \\ \textbf{85.12}  \\ \textbf{76.37} \\  \textbf{64$\pm$3} \end{tabular} &  
\begin{tabular}{@{}c@{}} \textbf{77.98} \\ \textbf{85.13} \\ \textbf{76.48} \\  \textbf{52$\pm$6} \end{tabular}  &  
\begin{tabular}{@{}c@{}} \textbf{90.61} \\ \textbf{87.98} \\ \textbf{93.44}  \\  \textbf{85$\pm$1} \end{tabular}&
\begin{tabular}{@{}c@{}} \textbf{71.01} \\ \textbf{39.31} \\  \textbf{91.33} \\  \textbf{46$\pm$1} \end{tabular}   \\
\hline

\begin{tabular}{@{}c@{}} ResNet \end{tabular} & \begin{tabular}{@{}c@{}} CelebA \\ CUB \\ S-ADD \\ CAD  \end{tabular}  &
\begin{tabular}{@{}c@{}}  80.83 \\ 82.77 \\ 74.7 \\ 55$\pm$5 \end{tabular}  & 
\begin{tabular}{@{}c@{}} 72.9 \\ 82.84 \\ 74.59 \\ 41$\pm$7 \end{tabular} &
 \begin{tabular}{@{}c@{}} 90.11 \\ 89.52 \\ 92.64 \\ 84$\pm$2 \end{tabular} & 
\begin{tabular}{@{}c@{}} 71.38 \\ 30.83 \\ 90.6  \\ 44$\pm$2  \end{tabular} & 
\begin{tabular}{@{}c@{}} 84.01 \\ 84.25 \\ 75.04 \\ 44$\pm$7  \end{tabular} & 
\begin{tabular}{@{}c@{}} 75.1 \\ 84.25 \\ 75.12 \\ 35$\pm$9  \end{tabular} &  
\begin{tabular}{@{}c@{}} 92.03 \\ 89.9 \\ 91.98 \\ 79$\pm$3  \end{tabular} & 
\begin{tabular}{@{}c@{}} 72.31 \\ 32.47 \\ 90.1 \\ 38$\pm$4 \end{tabular} & 
\begin{tabular}{@{}c@{}} \textbf{88.63} \\ \textbf{89.32} \\ \textbf{79.27} \\ \textbf{67$\pm$3}   \end{tabular} &  
\begin{tabular}{@{}c@{}} \textbf{81.1} \\ \textbf{89.45} \\ \textbf{79.39} \\ \textbf{55$\pm$6} \end{tabular}  &  
\begin{tabular}{@{}c@{}} \textbf{93.33} \\ \textbf{92.05} \\ \textbf{94.66} \\ \textbf{89$\pm$1} \end{tabular}&
\begin{tabular}{@{}c@{}} \textbf{74.68} \\ \textbf{43.33} \\ \textbf{93.19} \\ \textbf{50$\pm$1}  \end{tabular}   \\
\hline

\begin{tabular}{@{}c@{}} DenseNet \end{tabular} & \begin{tabular}{@{}c@{}} CelebA \\ CUB \\ S-ADD \\ CAD  \end{tabular}  &
\begin{tabular}{@{}c@{}}  80.07 \\ 80.62 \\ 73.25 \\  53$\pm$5 \end{tabular}  & 
\begin{tabular}{@{}c@{}} 72.2 \\ 80.66 \\ 73.12  \\ 39$\pm$7 \end{tabular} &
 \begin{tabular}{@{}c@{}} 89.87 \\ 87.57 \\ 91.26  \\ 82$\pm$2 \end{tabular} & 
\begin{tabular}{@{}c@{}} 70.24 \\ 29.13 \\ 89.6  \\ 42$\pm$2  \end{tabular} & 
\begin{tabular}{@{}c@{}} 82.75 \\ 82.02 \\ 73.66  \\ 42$\pm$7  \end{tabular} & 
\begin{tabular}{@{}c@{}} 74.1 \\ 82.46 \\ 73.74  \\ 33$\pm$9 \end{tabular} &  
\begin{tabular}{@{}c@{}} 91.7 \\  87.99 \\ 90.66  \\ 77$\pm$3 \end{tabular} & 
\begin{tabular}{@{}c@{}} 71.04 \\ 30.87 \\ 88.93  \\ 36$\pm$4 \end{tabular} & 
\begin{tabular}{@{}c@{}} \textbf{87.13} \\ \textbf{87.24} \\ \textbf{77.89}  \\ \textbf{65$\pm$3} \end{tabular} &  
\begin{tabular}{@{}c@{}} \textbf{80.01} \\ \textbf{87.25} \\ \textbf{78.01}  \\ \textbf{53$\pm$6} \end{tabular}  &  
\begin{tabular}{@{}c@{}} \textbf{92.99} \\ \textbf{90.01} \\ \textbf{93.77}  \\ \textbf{87$\pm$1} \end{tabular}&
\begin{tabular}{@{}c@{}} \textbf{73.31} \\ \textbf{41.53} \\ \textbf{92.44}  \\ \textbf{48$\pm$1} \end{tabular}   \\
\hline
\end{tabular}
}
\caption{Performance evaluation (in \%) on various databases by three widely used baseline networks; 'Acc' denotes accuracy; 'C MT' is the multi-task setting with coupling; 'NC MT' is the multi-task setting without coupling}
\label{comparison_celeba}
\end{table*}

\begin{table*}[t]
\centering
\scalebox{.9}{
\begin{tabular}{ |c||c|c|c||c|c|c|c| }
 \hline
\multicolumn{1}{|c||}{\begin{tabular}{@{}c@{}}   \end{tabular}}  &  \multicolumn{1}{c|}{\begin{tabular}{@{}c@{}}  NTS-ST \end{tabular}} & \multicolumn{1}{c|}{\begin{tabular}{@{}c@{}} NTS-NC MT \end{tabular}} & 
\multicolumn{1}{c||}{\begin{tabular}{@{}c@{}}  NTS-C MT\end{tabular}}  &
\multicolumn{1}{c|}{\begin{tabular}{@{}c@{}}  MGDA-UB\end{tabular}} 
&
\multicolumn{1}{c|}{\begin{tabular}{@{}c@{}}  GradNorm\end{tabular}} 
&
\multicolumn{1}{c|}{\begin{tabular}{@{}c@{}}  AdaShare\end{tabular}}
&
\multicolumn{1}{c|}{\begin{tabular}{@{}c@{}}  TR\end{tabular}}
\\
 \hline
 \hline

CUB   &  87.5 - 92  &  89.6 - 92.7  
& 94.4 - 95.2  & 86.3 - 90.9 & 86 - 90.4 & 86.2 - 90.6  &  83.23 - 76.5 \\
\hline
\hline

\begin{tabular}{@{}c@{}} S-ADD \end{tabular}  &   77.7 - 93 &  78 - 92.1 
&  82.6 - 94.9   & 76.5 - 93.1 &  76.1 - 92.8  & 76.7 - 93.4  & 73.1  - 78.67  \\
\hline
\hline

\begin{tabular}{@{}c@{}}  CAD \end{tabular}  &   60$\pm$5 - 86$\pm$2 &  49$\pm$6 - 81$\pm$4 
&   64$\pm$2 - 92$\pm$1 & 52$\pm$4 - 83$\pm$2   &  51$\pm$4  - 82$\pm$3 &  49$\pm$4 - 79$\pm$3  & 61$\pm$5  - 74$\pm$3  \\

\hline
\end{tabular}
}
\caption{Accuracy evaluation (in \%; in form Classes-Attributes) on various databases vs sota and MTL methods} 
\label{comparison_sota_2}
\end{table*}

\paragraph{Ablation Study}

Here we perform an ablation study, utilizing the Aff-Wild2, on the effect of each proposed coupling loss on the performance of TMIF-FEA. Table \ref{comparison_losses} shows the results when task relatedness was drawn from domain knowledge, or from the training set of Aff-Wild2. It can be seen that when TMIF-FEA was trained with either or both coupling losses under any relatedness scenario, its performance was superior to the case when no coupling loss has been used. Finally,  in both relatedness scenarios, best results have been achieved when TMIF-FEA was trained with both soft co-annotation and distr-matching losses. 
Similar results are yielded when we utilize each of the rest state-of-the-art on other databases, as explained in the previous subsection.

\subsection{Results on Case II: Beyond Affective Computing }  

\paragraph{Effectiveness of proposed coupling losses across broadly used networks}
Here, we utilized VGG-16, ResNet-50 and DenseNet-121. At first, we trained each of these networks for each application scenario for single-task learning (independent learning of classes and attributes) on CelebA, CUB, S-ADD and CAD datasets. These are denoted as '(2 $\times$) ST' in Table \ref{comparison_celeba}. We further trained these networks in MTL setting in two different cases: with coupling and without coupling during training. The former is denoted as 'MT-C'; the later as 'MT-NC' in Table \ref{comparison_celeba}. 
The presented results show the effectiveness of the proposed coupling losses to: i) avoid strong or mild negative transfer; ii) boost the performance of the multi-task models. The proposed coupling losses are network agnostic, as they bring similar level of improvement in all  utilized networks, tasks and databases.

\noindent
\underline{Face Recognition \& Fine-Grained Species Categorization}
Table \ref{comparison_celeba} shows that when the MTL baselines were trained without coupling, they displayed a better performance than the 2 single-task networks; this occurred in all studied cases, tasks, metrics and baseline models. This shows that the studied tasks were coherently correlated; training the multi-task architecture therefore, led to improved performance and no negative transfer occurred. 
Table \ref{comparison_celeba} further shows that when the baselines were trained in the multi-task setting with coupling, they greatly outperformed its counterpart  trained without coupling, in all studied tasks and metrics and for all baseline models. 
More precisely, when training with coupling, performance increased by 4.35\% and 6\% in Accuracy and F1 Score for identity classification, by  5.1\% and 5\% in  Accuracy and F1 Score for species categorization; and 1.25\% and 2.3\% in  Accuracy and F1 Score for facial attribute detection and 2.2\% and 10.8\% in  Accuracy and F1 Score for species attribute detection. 

\noindent
\underline{Shoe Type Recognition}
Table \ref{comparison_celeba} shows that negative transfer occurs in the case of attribute detection. Each single-task baseline for attribute detection displayed a better performance 
than its multi-task counterpart without coupling, whereas the latter displayed a better performance for shoe type classification. 
When the multi-task baseline networks were trained with coupling,  the performance on both tasks was boosted and outperformed single- and multi-task counterparts, by 4.2\% and 4.3\% for classification and 2.4\% and 2.9\% for detection in both metrics. The proposed coupling losses overcame the negative transfer that had occurred.

\noindent
\underline{Clothing Categories Recognition}
Table \ref{comparison_celeba} presents the outcomes of the 2-fold cross validation experiments (performed 6 times) in which the results are averaged and their spread is also shown (in the form: mean $\pm$ spread).
From Table \ref{comparison_celeba}, it can be seen that the selected tasks are very heterogeneous and less correlated as all multi-task baselines without coupling performed significantly worse than single-task counterparts in all utilized metrics. 
Such severe negative transfers occurs as   there is a big discrepancy in the size of labeled data per task in CAD dataset (the missing values for each attribute range from 12\% to 84\%) and its size is very small (it contains only 1856 images).
When the multi-task baselines were trained with coupling, negative transfer was prevented and the models significantly outperformed their single-task counterparts (10-14\% difference in Total Accuracy and 13-15\% in F1 Score for classification; 4-5\% in Total Accuracy and 5-7\% in F1 Score for attributes). Finally, a smaller spread of the results can be observed in the case when the models were trained with coupling.

\paragraph{Effectiveness of proposed coupling losses across the state-of-the-art}
At first, we show that the proposed coupling losses can also be incorporated in sota networks and thus we implement \textit{NTS-Net} \cite{yang2018learning} in single task setting (denoted NTS-ST), in MTL setting without coupling (NTS-MT NC) and in MTL setting by adding our proposed coupling losses (NTS-MT C).
Results are shown on Table \ref{comparison_sota_2} and are in accordance with the previous presented results (similar performance gain and alleviation of negative transfer).  
We then compare our method against MTL ones -presented in related work section- and thus we implement (ResNet50): MGDA-UB, GradNorm, AdaShare and TR.
Table \ref{comparison_sota_2} presents their results. Comparing these with ResNet50 C MT of Table \ref{comparison_celeba}, it is evident that our method significantly outperforms all of them.
Also, when comparing them to ST ResNet of Table \ref{comparison_celeba}: i) MGDA-UB, GradNorm and AdaShare cannot alleviate negative transfer in CAD for both tasks; ii) TR cannot alleviate negative transfer in CUB and CAD for attribute detection and in S-ADD for both tasks.

\section{Conclusion \& Limitation}
We proposed a method for accommodating  classification tasks by encoding prior knowledge of their relatedness. Our method is important as deep neural networks cannot necessarily capture tasks' relationship, especially in cases where: i) there is no or partial annotation overlap between tasks; ii) not many training data exist; iii) one task dominates the training process; iv) sub-optimal models for some tasks are produced; vi) there is big discrepancy in the size of labeled data per task. 
We considered a plethora of application scenarios and conducted extensive experimental studies. 
In all experiments our method helped the MT models greatly improve their performance compared to ST and MT models without coupling.
Our method further helped alleviate mild or significant negative transfer that occurred when MT models displayed worse performance in some or all studied tasks than ST models.
Our approach is general and flexible as long as there is a direct relationship between the studied tasks; the latter is our method's requirement and thus its limitation.

\bibliography{aaai24}

\begin{thebibliography}{49}
\providecommand{\natexlab}[1]{#1}

\bibitem[{Benitez-Quiroz, Srinivasan, and Martinez(2016)}]{EmotioNet2016}
Benitez-Quiroz, C.; Srinivasan, R.; and Martinez, A. 2016.
\newblock EmotioNet: An accurate, real-time algorithm for the automatic annotation of a million facial expressions in the wild.
\newblock In \emph{Proceedings of IEEE International Conference on Computer Vision \& Pattern Recognition (CVPR'16)}. Las Vegas, NV, USA.

\bibitem[{Chen, Gallagher, and Girod(2012)}]{chen2012describing}
Chen, H.; Gallagher, A.; and Girod, B. 2012.
\newblock Describing clothing by semantic attributes.
\newblock In \emph{European conference on computer vision}, 609--623. Springer.

\bibitem[{Chen et~al.(2018)Chen, Badrinarayanan, Lee, and Rabinovich}]{chen2018gradnorm}
Chen, Z.; Badrinarayanan, V.; Lee, C.-Y.; and Rabinovich, A. 2018.
\newblock Gradnorm: Gradient normalization for adaptive loss balancing in deep multitask networks.
\newblock In \emph{International Conference on Machine Learning}, 794--803. PMLR.

\bibitem[{Deng, Chen, and Shi(2020)}]{deng2020multitask}
Deng, D.; Chen, Z.; and Shi, B.~E. 2020.
\newblock Multitask emotion recognition with incomplete labels.
\newblock In \emph{2020 15th IEEE International Conference on Automatic Face and Gesture Recognition (FG 2020)(FG)}, 828--835. IEEE Computer Society.

\bibitem[{Deng et~al.(2020)Deng, Guo, Ververas, Kotsia, and Zafeiriou}]{deng2020retinaface}
Deng, J.; Guo, J.; Ververas, E.; Kotsia, I.; and Zafeiriou, S. 2020.
\newblock Retinaface: Single-shot multi-level face localisation in the wild.
\newblock In \emph{Proceedings of the IEEE/CVF Conference on Computer Vision and Pattern Recognition}, 5203--5212.

\bibitem[{Du, Tao, and Martinez(2014)}]{du2014compound}
Du, S.; Tao, Y.; and Martinez, A.~M. 2014.
\newblock Compound facial expressions of emotion.
\newblock \emph{Proceedings of the National Academy of Sciences}, 111(15): E1454--E1462.

\bibitem[{Ekman(1997)}]{ekman1997face}
Ekman, R. 1997.
\newblock \emph{What the face reveals: Basic and applied studies of spontaneous expression using the Facial Action Coding System (FACS)}.
\newblock Oxford University Press, USA.

\bibitem[{Fabian Benitez-Quiroz, Srinivasan, and Martinez(2016)}]{fabian2016EmotioNet}
Fabian Benitez-Quiroz, C.; Srinivasan, R.; and Martinez, A.~M. 2016.
\newblock Emotionet: An accurate, real-time algorithm for the automatic annotation of a million facial expressions in the wild.
\newblock In \emph{Proceedings of the IEEE Conference on Computer Vision and Pattern Recognition}, 5562--5570.

\bibitem[{Hinton, Vinyals, and Dean(2015)}]{hinton2015distilling}
Hinton, G.; Vinyals, O.; and Dean, J. 2015.
\newblock Distilling the knowledge in a neural network.
\newblock \emph{arXiv:1503.02531}.

\bibitem[{Jeong et~al.(2022)Jeong, Hong, Oh, Hong, Jeong, and Jung}]{jeong2022learning}
Jeong, J.-Y.; Hong, Y.-G.; Oh, J.; Hong, S.; Jeong, J.-W.; and Jung, Y. 2022.
\newblock Learning from Synthetic Data: Facial Expression Classification based on Ensemble of Multi-task Networks.
\newblock \emph{arXiv preprint arXiv:2207.10025}.

\bibitem[{Khorrami, Paine, and Huang(2015)}]{khorrami2015deep}
Khorrami, P.; Paine, T.; and Huang, T. 2015.
\newblock Do deep neural networks learn facial action units when doing expression recognition?
\newblock In \emph{Proceedings of the IEEE International Conference on Computer Vision Workshops}, 19--27.

\bibitem[{Kokkinos(2017)}]{kokkinos2017ubernet}
Kokkinos, I. 2017.
\newblock Ubernet: Training a universal convolutional neural network for low-, mid-, and high-level vision using diverse datasets and limited memory.
\newblock In \emph{Proceedings of the IEEE Conference on Computer Vision and Pattern Recognition}, 6129--6138.

\bibitem[{Kollias(2022{\natexlab{a}})}]{kollias2022eccv}
Kollias, D. 2022{\natexlab{a}}.
\newblock ABAW: learning from synthetic data \& multi-task learning challenges.
\newblock In \emph{European Conference on Computer Vision}, 157--172. Springer.

\bibitem[{Kollias(2022{\natexlab{b}})}]{kollias2022cvpr}
Kollias, D. 2022{\natexlab{b}}.
\newblock Abaw: Valence-arousal estimation, expression recognition, action unit detection \& multi-task learning challenges.
\newblock In \emph{Proceedings of the IEEE/CVF Conference on Computer Vision and Pattern Recognition}, 2328--2336.

\bibitem[{Kollias et~al.(2020{\natexlab{a}})Kollias, Cheng, Ververas, Kotsia, and Zafeiriou}]{kollias2020deep}
Kollias, D.; Cheng, S.; Ververas, E.; Kotsia, I.; and Zafeiriou, S. 2020{\natexlab{a}}.
\newblock Deep neural network augmentation: Generating faces for affect analysis.
\newblock \emph{International Journal of Computer Vision}, 128(5): 1455--1484.

\bibitem[{Kollias et~al.(2017)Kollias, Nicolaou, Kotsia, Zhao, and Zafeiriou}]{kollias2017recognition}
Kollias, D.; Nicolaou, M.~A.; Kotsia, I.; Zhao, G.; and Zafeiriou, S. 2017.
\newblock Recognition of affect in the wild using deep neural networks.
\newblock In \emph{Proceedings of the IEEE Conference on Computer Vision and Pattern Recognition Workshops}, 26--33.

\bibitem[{Kollias et~al.(2020{\natexlab{b}})Kollias, Schulc, Hajiyev, and Zafeiriou}]{kollias2020analysing}
Kollias, D.; Schulc, A.; Hajiyev, E.; and Zafeiriou, S. 2020{\natexlab{b}}.
\newblock Analysing affective behavior in the first abaw 2020 competition.
\newblock In \emph{2020 15th IEEE International Conference on Automatic Face and Gesture Recognition (FG 2020)}, 637--643. IEEE.

\bibitem[{Kollias et~al.(2023)Kollias, Tzirakis, Baird, Cowen, and Zafeiriou}]{kollias2023cvpr}
Kollias, D.; Tzirakis, P.; Baird, A.; Cowen, A.; and Zafeiriou, S. 2023.
\newblock Abaw: Valence-arousal estimation, expression recognition, action unit detection \& emotional reaction intensity estimation challenges.
\newblock In \emph{Proceedings of the IEEE/CVF Conference on Computer Vision and Pattern Recognition}, 5888--5897.

\bibitem[{Kollias et~al.(2019)Kollias, Tzirakis, Nicolaou, Papaioannou, Zhao, Schuller, Kotsia, and Zafeiriou}]{kollias2019deep}
Kollias, D.; Tzirakis, P.; Nicolaou, M.~A.; Papaioannou, A.; Zhao, G.; Schuller, B.; Kotsia, I.; and Zafeiriou, S. 2019.
\newblock Deep affect prediction in-the-wild: Aff-wild database and challenge, deep architectures, and beyond.
\newblock \emph{International Journal of Computer Vision}, 1--23.

\bibitem[{Kollias and Zafeiriou(2019)}]{kollias2019expression}
Kollias, D.; and Zafeiriou, S. 2019.
\newblock Expression, Affect, Action Unit Recognition: Aff-Wild2, Multi-Task Learning and ArcFace.
\newblock \emph{arXiv preprint arXiv:1910.04855}.

\bibitem[{Kollias and Zafeiriou(2021{\natexlab{a}})}]{kollias2021affect}
Kollias, D.; and Zafeiriou, S. 2021{\natexlab{a}}.
\newblock Affect Analysis in-the-wild: Valence-Arousal, Expressions, Action Units and a Unified Framework.
\newblock \emph{arXiv preprint arXiv:2103.15792}.

\bibitem[{Kollias and Zafeiriou(2021{\natexlab{b}})}]{kollias2021analysing}
Kollias, D.; and Zafeiriou, S. 2021{\natexlab{b}}.
\newblock Analysing affective behavior in the second abaw2 competition.
\newblock In \emph{Proceedings of the IEEE/CVF International Conference on Computer Vision}, 3652--3660.

\bibitem[{Kovashka, Parikh, and Grauman(2012)}]{kovashka2012whittlesearch}
Kovashka, A.; Parikh, D.; and Grauman, K. 2012.
\newblock Whittlesearch: Image search with relative attribute feedback.
\newblock In \emph{2012 IEEE Conference on Computer Vision and Pattern Recognition}, 2973--2980. IEEE.

\bibitem[{Li, Deng, and Du(2017)}]{li2017reliable}
Li, S.; Deng, W.; and Du, J. 2017.
\newblock Reliable crowdsourcing and deep locality-preserving learning for expression recognition in the wild.
\newblock In \emph{Proceedings of the IEEE Conference on Computer Vision and Pattern Recognition}, 2852--2861.

\bibitem[{Lin et~al.(2019)Lin, Zhen, Li, Zhang, and Kwong}]{lin2019pareto}
Lin, X.; Zhen, H.-L.; Li, Z.; Zhang, Q.; and Kwong, S. 2019.
\newblock Pareto Multi-Task Learning.
\newblock In \emph{Thirty-third Conference on Neural Information Processing Systems (NeurIPS 2019)}.

\bibitem[{Liu, Liang, and Gitter(2019)}]{liu2019loss}
Liu, S.; Liang, Y.; and Gitter, A. 2019.
\newblock Loss-balanced task weighting to reduce negative transfer in multi-task learning.
\newblock In \emph{Proceedings of the AAAI Conference on Artificial Intelligence}, volume~33, 9977--9978.

\bibitem[{Liu et~al.(2015)Liu, Luo, Wang, and Tang}]{liu2015faceattributes}
Liu, Z.; Luo, P.; Wang, X.; and Tang, X. 2015.
\newblock Deep Learning Face Attributes in the Wild.
\newblock In \emph{Proceedings of International Conference on Computer Vision (ICCV)}.

\bibitem[{Mollahosseini, Hasani, and Mahoor(2017)}]{mollahosseini2017affectnet}
Mollahosseini, A.; Hasani, B.; and Mahoor, M.~H. 2017.
\newblock Affectnet: A database for facial expression, valence, and arousal computing in the wild.
\newblock \emph{arXiv preprint arXiv:1708.03985}.

\bibitem[{Psaroudakis and Kollias(2022)}]{psaroudakis2022mixaugment}
Psaroudakis, A.; and Kollias, D. 2022.
\newblock Mixaugment \& mixup: Augmentation methods for facial expression recognition.
\newblock In \emph{Proceedings of the IEEE/CVF Conference on Computer Vision and Pattern Recognition}, 2367--2375.

\bibitem[{Ranjan et~al.(2017)Ranjan, Sankaranarayanan, Castillo, and Chellappa}]{ranjan2017all}
Ranjan, R.; Sankaranarayanan, S.; Castillo, C.~D.; and Chellappa, R. 2017.
\newblock An all-in-one convolutional neural network for face analysis.
\newblock In \emph{2017 12th IEEE International Conference on Automatic Face \& Gesture Recognition (FG 2017)}, 17--24. IEEE.

\bibitem[{Ryumina, Dresvyanskiy, and Karpov(2022)}]{ryumina2022search}
Ryumina, E.; Dresvyanskiy, D.; and Karpov, A. 2022.
\newblock In search of a robust facial expressions recognition model: A large-scale visual cross-corpus study.
\newblock \emph{Neurocomputing}, 514: 435--450.

\bibitem[{Savchenko(2021)}]{savchenko2021facial}
Savchenko, A.~V. 2021.
\newblock Facial expression and attributes recognition based on multi-task learning of lightweight neural networks.
\newblock In \emph{2021 IEEE 19th International Symposium on Intelligent Systems and Informatics (SISY)}, 119--124. IEEE.

\bibitem[{Savchenko(2022)}]{savchenko2022hse}
Savchenko, A.~V. 2022.
\newblock HSE-NN Team at the 4th ABAW Competition: Multi-task Emotion Recognition and Learning from Synthetic Images.
\newblock \emph{arXiv preprint arXiv:2207.09508}.

\bibitem[{Sener and Koltun(2018)}]{sener2018multi}
Sener, O.; and Koltun, V. 2018.
\newblock Multi-task learning as multi-objective optimization.
\newblock \emph{arXiv preprint arXiv:1810.04650}.

\bibitem[{Strezoski, Noord, and Worring(2019)}]{strezoski2019many}
Strezoski, G.; Noord, N.~v.; and Worring, M. 2019.
\newblock Many task learning with task routing.
\newblock In \emph{Proceedings of the IEEE/CVF International Conference on Computer Vision}, 1375--1384.

\bibitem[{Sun et~al.(2020)Sun, Panda, Feris, and Saenko}]{sun2020adashare}
Sun, X.; Panda, R.; Feris, R.; and Saenko, K. 2020.
\newblock Adashare: Learning what to share for efficient deep multi-task learning.
\newblock \emph{Advances in Neural Information Processing Systems}, 33: 8728--8740.

\bibitem[{Vo et~al.(2020)Vo, Lee, Yang, and Kim}]{vo2020pyramid}
Vo, T.-H.; Lee, G.-S.; Yang, H.-J.; and Kim, S.-H. 2020.
\newblock Pyramid with super resolution for in-the-wild facial expression recognition.
\newblock \emph{IEEE Access}, 8: 131988--132001.

\bibitem[{Wah et~al.(2011)Wah, Branson, Welinder, Perona, and Belongie}]{WahCUB_200_2011}
Wah, C.; Branson, S.; Welinder, P.; Perona, P.; and Belongie, S. 2011.
\newblock {The Caltech-UCSD Birds-200-2011 Dataset}.
\newblock Technical Report CNS-TR-2011-001, California Institute of Technology.

\bibitem[{Wang, Fidler, and Urtasun(2015)}]{wang2015holistic}
Wang, S.; Fidler, S.; and Urtasun, R. 2015.
\newblock Holistic 3d scene understanding from a single geo-tagged image.
\newblock In \emph{Proceedings of the IEEE Conference on Computer Vision and Pattern Recognition}, 3964--3972.

\bibitem[{Wang, Gan, and Ji(2017)}]{wang2017expression}
Wang, S.; Gan, Q.; and Ji, Q. 2017.
\newblock Expression-assisted facial action unit recognition under incomplete AU annotation.
\newblock \emph{Pattern Recognition}, 61: 78--91.

\bibitem[{Wang et~al.(2019)Wang, Dai, P{\'o}czos, and Carbonell}]{wang2019characterizing}
Wang, Z.; Dai, Z.; P{\'o}czos, B.; and Carbonell, J. 2019.
\newblock Characterizing and avoiding negative transfer.
\newblock In \emph{Proceedings of the IEEE/CVF Conference on Computer Vision and Pattern Recognition}, 11293--11302.

\bibitem[{Wang et~al.(2017)Wang, He, Fu, Feng, Jiang, and Xue}]{wang2017multi}
Wang, Z.; He, K.; Fu, Y.; Feng, R.; Jiang, Y.-G.; and Xue, X. 2017.
\newblock Multi-task deep neural network for joint face recognition and facial attribute prediction.
\newblock In \emph{Proceedings of the 2017 ACM on International Conference on Multimedia Retrieval}, 365--374. ACM.

\bibitem[{Wu, Zhang, and R{\'e}(2019)}]{wu2019understanding}
Wu, S.; Zhang, H.~R.; and R{\'e}, C. 2019.
\newblock Understanding and Improving Information Transfer in Multi-Task Learning.
\newblock In \emph{International Conference on Learning Representations}.

\bibitem[{Yang et~al.(2016)Yang, Wu, Wang, and Ji}]{yang2016multiple}
Yang, J.; Wu, S.; Wang, S.; and Ji, Q. 2016.
\newblock Multiple facial action unit recognition enhanced by facial expressions.
\newblock In \emph{2016 23rd International Conference on Pattern Recognition (ICPR)}, 4089--4094. IEEE.

\bibitem[{Yang et~al.(2018)Yang, Luo, Wang, Hu, Gao, and Wang}]{yang2018learning}
Yang, Z.; Luo, T.; Wang, D.; Hu, Z.; Gao, J.; and Wang, L. 2018.
\newblock Learning to navigate for fine-grained classification.
\newblock In \emph{Proceedings of the European conference on computer vision (ECCV)}, 420--435.

\bibitem[{Yu et~al.(2020)Yu, Kumar, Gupta, Levine, Hausman, and Finn}]{yu2020gradient}
Yu, T.; Kumar, S.; Gupta, A.; Levine, S.; Hausman, K.; and Finn, C. 2020.
\newblock Gradient Surgery for Multi-Task Learning.
\newblock \emph{Advances in Neural Information Processing Systems}, 33.

\bibitem[{Zafeiriou et~al.(2017)Zafeiriou, Kollias, Nicolaou, Papaioannou, Zhao, and Kotsia}]{zafeiriou2017aff}
Zafeiriou, S.; Kollias, D.; Nicolaou, M.~A.; Papaioannou, A.; Zhao, G.; and Kotsia, I. 2017.
\newblock Aff-wild: Valence and arousal ‘in-the-wild’challenge.
\newblock In \emph{Computer Vision and Pattern Recognition Workshops (CVPRW), 2017 IEEE Conference on}, 1980--1987. IEEE.

\bibitem[{Zamir et~al.(2018)Zamir, Sax, Shen, Guibas, Malik, and Savarese}]{zamir2018taskonomy}
Zamir, A.~R.; Sax, A.; Shen, W.; Guibas, L.~J.; Malik, J.; and Savarese, S. 2018.
\newblock Taskonomy: Disentangling task transfer learning.
\newblock In \emph{Proceedings of the IEEE Conference on Computer Vision and Pattern Recognition}, 3712--3722.

\bibitem[{Zhang et~al.(2022)Zhang, Qiu, Wang, Zeng, Zhang, An, Ma, and Ding}]{zhang2022transformer}
Zhang, W.; Qiu, F.; Wang, S.; Zeng, H.; Zhang, Z.; An, R.; Ma, B.; and Ding, Y. 2022.
\newblock Transformer-based Multimodal Information Fusion for Facial Expression Analysis.
\newblock In \emph{Proceedings of the IEEE/CVF Conference on Computer Vision and Pattern Recognition}, 2428--2437.

\end{thebibliography}

\end{document}